\documentclass[letterpaper, 10 pt, conference]{ieeeconf}  % Comment this line out if you need a4paper

\IEEEoverridecommandlockouts                              % This command is only needed if 
\usepackage{graphics} % for pdf, bitmapped graphics files
\usepackage{epsfig} % for postscript graphics files
\usepackage{xcolor}
\usepackage{amsmath,amssymb,amsfonts}
\usepackage[english]{babel}
\usepackage{caption}
\usepackage{subcaption}
\usepackage{comment}
\usepackage{url}
\usepackage{bm}
\usepackage{multirow}
\usepackage[noadjust]{cite}
\usepackage{hyperref}
\hypersetup{
    colorlinks=true,
    linkcolor=black,
    urlcolor=blue,
}

\title{\LARGE \bf
The Reasonable Crowd: Towards evidence-based and interpretable models of driving behavior
}

\author{Bassam Helou$^{1}$, Aditya Dusi$^{1}$, Anne Collin$^{1}$, Noushin Mehdipour$^{1}$, Zhiliang Chen$^{1}$, Cristhian Lizarazo$^{1}$, \\ Calin Belta$^{1,2}$, Tichakorn Wongpiromsarn$^{3}$, Radboud Duintjer Tebbens$^{1}$, Oscar Beijbom.$^{1}$ % 
\thanks{$^{1}$ Motional. Boston, MA.}%
\thanks{$^{2}$ Boston University, Boston, MA.}
\thanks{$^{3}$ Iowa State University, IA.}%
}

\begin{document}

\maketitle
\thispagestyle{empty}
\pagestyle{empty}

%%%%%%%%%%%%%%%%%%%%%%%%%%%%%%%%%%%%%%%%%%%%%%%%%%%%%%%%%%%%%%%%%%%%%%%%%%%%%%%%
\begin{abstract}
Autonomous vehicles must balance a complex set of objectives. There is no consensus on how they should do so, nor on a model for specifying a desired driving behavior. We created a dataset to help address some of these questions in a limited operating domain. 
The data consists of 92 traffic scenarios, with multiple ways of traversing each scenario. Multiple annotators expressed their preference between pairs of scenario traversals.
We used the data to compare an instance of a rulebook \cite{Censi2020}, carefully hand-crafted independently of the dataset, with several interpretable machine learning models such as Bayesian networks, decision trees, and logistic regression trained on the dataset. To compare driving behavior, these models use scores indicating by how much different scenario traversals violate each of 14 driving rules. The rules are interpretable and designed by subject-matter experts.
First, we found that these rules were enough for these models to achieve a high classification accuracy on the dataset. Second, we found that the rulebook provides high interpretability without excessively sacrificing performance. Third, the data pointed to possible improvements in the rulebook and the rules, and to potential new rules. Fourth, we explored the interpretability vs performance trade-off by also training non-interpretable models such as a random forest. Finally, we make the dataset publicly available to encourage a discussion from the wider community on behavior specification for AVs. Please find it at \url{github.com/bassam-motional/Reasonable-Crowd}.
\end{abstract}

%%%%%%%%%%%%%%%%%%%%%%%%%%%%%%%%%%%%%%%%%%%%%%%%%%%%%%%%%%%%%%%%%%%%%%%%%%%%%%%%
\section{Introduction}
Designing and testing a safe autonomous vehicle (AV) critically depends on a system-level specification of the desired driving behavior \cite{Collin2020}. Determining such a specification is difficult. AVs balance a complex set of objectives, such as driving safely and comfortably, complying with numerous traffic laws, getting to a destination, and generally meeting ethical and cultural expectations of reasonable driving behavior \cite{nolte2017towards,shalev2017formal,parseh2019pre,ulbrich2013probabilistic,qian2014priority,iso2019pas,moral_machine,Censi2020}. There is no consensus on a driving behavior specification that balances these considerations, nor does there exist a standard framework for such a specification. 

We expect a model for specifying driving behavior to be both interpretable and evidence-based. Indeed, the US National Highway Traffic Safety Administration (NHTSA) encourages AV developers ``to have a documented process for the assessment, testing, and validation of a variety of behavioral competencies [including] obeying traffic laws, following reasonable road etiquette, and responding to other vehicles or hazards.'' % RT Please cite  https://www.nhtsa.gov/sites/nhtsa.dot.gov/files/documents/13069a-ads2.0_090617_v9a_tag.pdf, p. 7

Interpretable models have been proposed to specify or validate the driving behavior of AVs \cite{Censi2020, Assume-Guarantee, cai2020rules, corso2020interpretable}. Central to this paper is the Rulebooks approach \cite{Censi2020}, which proposes a set of interpretable rules endowed with a priority structure to rank driving behaviors. The priority structure can be a pre-order \cite{Censi2020} or a total order over equivalence classes \cite{xiao2021rulebased}. We also examine interpretable Machine Learning (ML) models, including Bayesian networks, decision trees, and linear support vector machines, as well as non-interpretable ML models such as random forests and neural networks to explore the interpretability vs performance trade-off.

Currently, there does not exist publicly available preference data to validate, develop and compare models of driving behavior. We created a dataset consisting of annotators' preferences between two ways of navigating the same traffic situation. We then used this dataset to compare and contrast the Rulebooks approach with the ML methods enumerated above.
Furthermore, we make the dataset publicly available to encourage a public discussion on how AVs should behave and to stimulate research on driving behavior models. Please find it at \url{github.com/bassam-motional/Reasonable-Crowd}.

The main contributions of this paper are as follows: 
\begin{itemize}
    \item We compare different driving specification models, drawing from different fields, based on accuracy, interpretability, and ease of implementation.
    \item We use the data to gather insights on the Rulebooks approach proposed in \cite{Censi2020}. 
    \item We provide an annotated dataset to accelerate the development and testing of driving behavior specifications.
\end{itemize}

We organized the paper as follows. Sec. II reviews related work. Sec. III discusses the Reasonable Crowd dataset. Sec. IV presents the rulebooks model and the ML models we use in this study. In Sec. V, we evaluate the performance of these models against the dataset, and discuss the results.

\section{Related Work}

Two main types of data have been used to help determine reward or cost functions that set the desired behavior of an autonomous agent. The first is {\em human demonstrations}, which are popular in Inverse Reinforcement Learning. Demonstrations are used in \cite{max_margin_uber, apprecntiship_learning, driving_styles} to learn a reward function that is a weighted sum of hand-designed features $\phi_j(\cdot)$:
\begin{equation}
\Phi\left(s\right)=\sum_{j=1}^{k}w_{j}\phi_{j}\left(s\right),
\label{eq:linear_reward}
\end{equation}
where the $w_j$ are weights learned from data. 
Gaussian processes are used in \cite{GPIRL} to capture complex reward structures. The authors of \cite{cho2019deep} use deep learning to determine a margin to the satisfaction of each rule based on context. Deep learning is also used in  \cite{wulfmeier2016maximum, watchthis, guided_cost_learning} to compute a reward function. By deviating from Eq. (\ref{eq:linear_reward}), these approaches sacrifice interpretability, but do not quantify the resulting performance gain. In this paper, we explore the trade-off between interpretability and performance.

The second type is {\em preference data}, where annotators indicate their preferences between trajectories. In most studies, the wisdom of the crowd, rather than experts, determines these preferences. Using the crowd has a long history \cite{ox}, and in ML many datasets and studies make use of it such as \cite{chalearn, protest}.
The authors of \cite{Sadigh2017ActivePL, biyik2018batch, coactive_feedback} use preference data to learn a reward or cost function similar to Eq. (\ref{eq:linear_reward}). A more general linear cost function that includes context and features with learnable parameters is proposed in \cite{PlanIt}. Our work is closest to \cite{Sadigh2017ActivePL, biyik2018batch}. However, we do not use active learning. Instead, we create more realistic data at a much larger scale, with more diverse traffic scenarios selected to highlight trade-offs between driving rules.

Both types of data have their drawbacks. As suggested in \cite{atari, preference_and_irl}, it is desirable to use both to determine behavior specifications. Human demonstrations are more realistic but also more difficult to obtain, especially for rare or dangerous traffic scenarios. With preference data obtained from simulations, it is possible to query different aspects of driving behavior in diverse traffic scenarios by constructing multiple trajectories for comparison in each scenario. However, simulation data has its limitations. For example, it is difficult for annotators to judge comfort and speed from a simulation.

Finally, while there is no consensus on how to best validate and compare behavior specifications for AVs, different strategies are emerging. The CARLA challenge is a public benchmark based on closed loop simulation \cite{Carla}. A closed-loop ML-based planning benchmark for autonomous driving from human demonstrations is under development  \cite{nuPlan}. Furthermore, some metrics can be estimated on large public perception datasets such as nuScenes \cite{nuscenes}. For example, in \cite{endtoend}, collision rates are estimated by determining when the planned trajectory will overlap with other vehicles in the future. Note that this study is based on a large private perception dataset. Our dataset could be another strategy to test behavior specifications. It has a lower barrier to entry, and can test more nuanced aspects of driving behavior such as road etiquette expectations related to clearance.

\section{The Reasonable Crowd Dataset}

We present the details of the dataset we use.

\subsection{Terminology}

We use the following terminology to describe the dataset:
\begin{itemize}
    \item Ego: the AV whose behavior we examine.
    \item Map: the topology of the road network and the static environment (e.g., buildings, road markings).
    \item Trajectory: An agent's path on the map as a function of time.
    \item Scenario: a map populated with other agents, their starting positions, and their trajectories.
    \item Realization: a scenario and ego’s trajectory through it.
    \item Operating domain: the driving and environment conditions ego encounters. %, as well as restrictions on the scenario.
\end{itemize}

\subsection{Data}

We created 576 realizations spread over 92 scenarios. The realizations are not representative of safe AV behavior. We only use the dataset to evaluate rules and different models of driving behavior.

We manually constructed some of the scenarios. To increase their diversity and complexity, we used an in-house traffic simulator to produce many candidate scenarios and realizations and then manually selected some of them.

We used an old version of an in-house 3D driving simulator to render the scenarios. These renderings add immersion and allow annotators to ingest complex scenarios. Nonetheless, our rendering quality has limitations. For example, the renderings do not always clearly show whether a collision occurred, which reduced the quality of our annotations.

\begin{figure}[tb]
  \centering
    \includegraphics[scale=0.35]{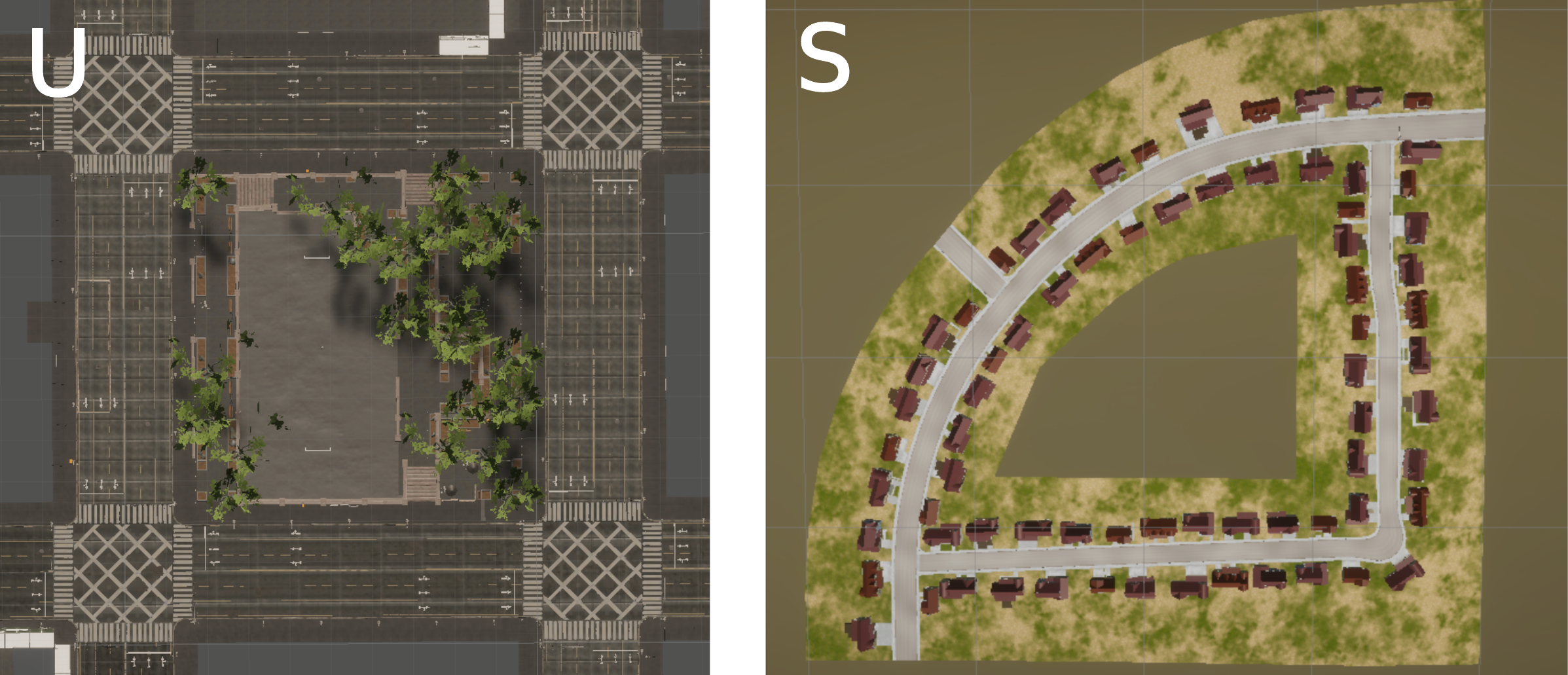}
  \caption{The maps used in this study. Map U is an urban environment with 4-way intersections. Map S is a small suburban environment. }
  \label{Maps}
\end{figure}

Our operating domain is limited to sunny weather in the two simple environments shown in Fig. \ref{Maps}. We do not include traffic lights, stop signs, bicycles or other personal mobility devices, or cars reversing.
Since the performance of the models we study depends on our data's distribution, we strove to create a dataset with an adequate coverage of our operating domain. Broader operating domains will necessitate principled approaches to ensure appropriate coverage of scenarios.
Finally, we note that for simplicity we fixed the agents' behavior. Agents do not react to ego's trajectory. 

\subsection{Annotations \label{anns}}

\begin{figure}[tb]
  \centering
    \includegraphics[scale=0.13]{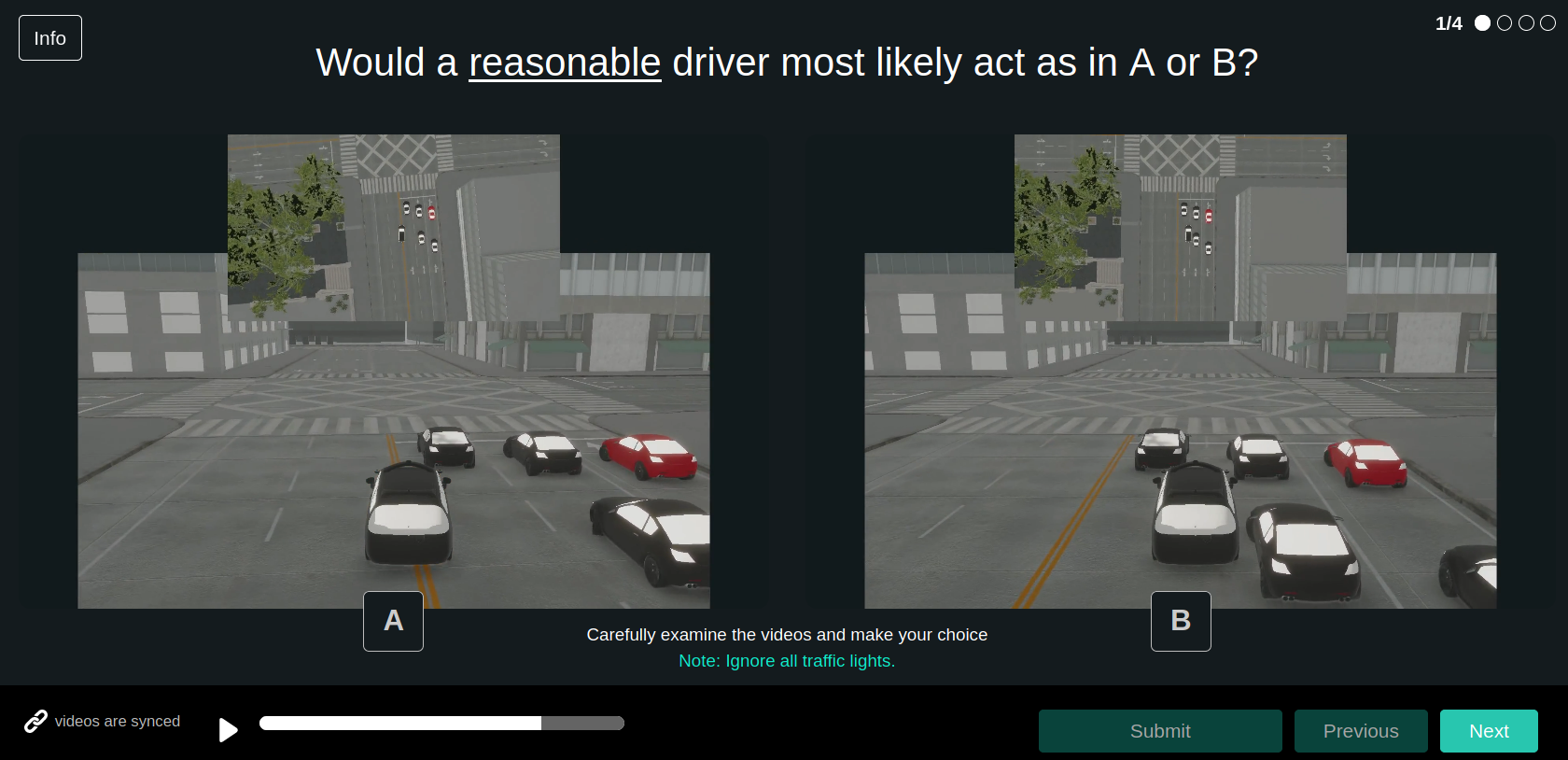}
  \caption{Our custom annotation tool with the instruction we gave to annotators. Ten out of 14 annotators picked option B in the example shown. }
  \label{ann_tool}
\end{figure}

 On Amazon Mechanical Turk, we showed annotators videos of ego traversing the {\it same} scenario in two different ways. We asked them to pick the video exhibiting more reasonable driving behavior, as shown in Fig. \ref{ann_tool}.
 Sixty-five annotators compared 1682 unique realization pairs. We collected 40560 annotations, with multiple different annotators annotating each realization pair. We did this to reduce noise and because preferences for driving behaviors vary across different humans - there is no objectively best way to drive. 

\begin{figure*}[h]
    \centering
    \includegraphics[scale=0.4]{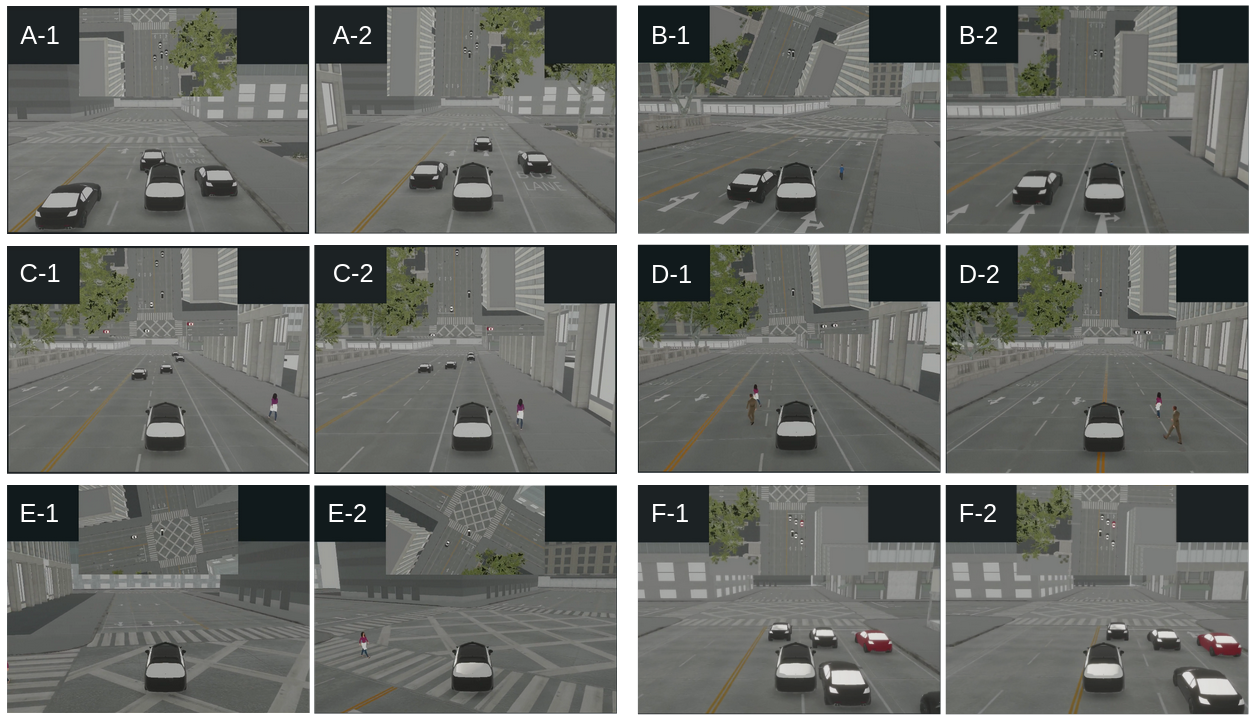}
    \caption{
    Example realization pairs from the dataset. The realizations can be seen in their entirety by clicking on the links in this caption. \textbf{A}: Trade-off of left vs right vehicle clearance. 11 out of 14 annotators chose \href{https://api-reasonable-crowd.ml.motional.com/v1/track/5532451e401543ac826f298439cdb921/video}{A-1} over \href{https://api-reasonable-crowd.ml.motional.com/v1/track/8ceab3955700473b9efc902b07404120/video}{A-2}.
\textbf{B}: Trade-off of a VRU collision vs vehicle collision. 29 out of 34 annotators chose \href{https://api-reasonable-crowd.ml.motional.com/v1/track/86dd2141c4c24b2281241e77f845b3a8/video}{B-1} over \href{https://api-reasonable-crowd.ml.motional.com/v1/track/f5a369ae6add4d699c6a80ba372b8098/video}{B-2}.
\textbf{C}: Trade-off of various degrees of pedestrian off road clearance violations. 0 out of 14 annotators chose \href{https://api-reasonable-crowd.ml.motional.com/v1/track/64034f44c03f4b7cb963b9866022f339/video}{C-1} over \href{https://api-reasonable-crowd.ml.motional.com/v1/track/e72dd04bc32e4d2ab91e269d7fbffd8b/video}{C-2}.
\textbf{D}: Trade-off of various degrees of pedestrian on road clearance violations. 33 out of 34 annotators chose \href{https://api-reasonable-crowd.ml.motional.com/v1/track/48f1d96333b24686815301fb02ffbf5f/video}{D-1} over \href{https://api-reasonable-crowd.ml.motional.com/v1/track/d9059863a9a34d1397adfbd71088dff8/video}{D-2}.
\textbf{E}: Examples of realizations with pedestrians crossing a crosswalk. 13 out of 20 annotators chose \href{https://api-reasonable-crowd.ml.motional.com/v1/track/46f03dafcc36426b972ebb519e0fda76/video}{E-1} over \href{https://api-reasonable-crowd.ml.motional.com/v1/track/364490e57f5d4a03882524392470f829/video}{E-2}.
\textbf{F}: Trade-off of a major violation of right vehicle clearance vs a minor violation of driving on the correct side of the road. 0 out of 14 annotators chose \href{https://api-reasonable-crowd.ml.motional.com/v1/track/1425aab3bd9d4c1aa3e109f0984dd578/video}{F-1} over \href{https://api-reasonable-crowd.ml.motional.com/v1/track/e6483c18e611460e839bbd66f5abfe81/video}{F-2}.
}
    \label{fig:qualitative_examples}
\end{figure*}

 To collect quality annotations, we recruit annotators with more than 3000 approved tasks and an approval rating $\ge$ 97\%. We require that they have more than 20000 miles of driving experience in North America (based on a questionnaire asking about transportation habits). They must also pass a test consisting of pairs that are easy to compare.
 Despite these qualifications, annotations can be noisy (see Fig. \ref{fig:qualitative_examples} for example annotations). This had limited impact on our analysis because when comparing realizations we aggregate annotations into the ground-truth labels of +1 and -1. 

Finally, we designed our dataset to contain realizations that are of varying difficulty to compare. When deployed, driving behavior frameworks will sometimes have to choose between imperfect trajectories that make difficult trade-offs. To get a sense of the variety in the difficulty of our dataset, Table \ref{ann_pair_agreement} shows the number of realization pairs stratified by the level of inter-annotator agreement 
\begin{equation}
\label{eq:agreement_level}
a\left(w_{1}, w_{2}\right) = \left|n_{w_{1}>w_{2}}-n_{w_{2}>w_{1}}\right|/\left(n_{w_{1}>w_{2}}+n_{w_{2}>w_{1}}\right),
\end{equation}
where $w_{1}$ and $w_{2}$ are two realizations, and $n_{w > w'}$ is the number of annotators that preferred realization $w$ over realization $w'$. The closer $a$ is to $0$, the less consensus there was among annotators regarding that pair, and so the more difficult the comparison.

\begin{table}[tb]
\caption{Distribution of realization pairs based on the inter-annotator agreement given by Eq. (\ref{eq:agreement_level})}.
\label{ann_pair_agreement}
\begin{center}

\begin{tabular}{|c|c|c|c|c|c|}
\hline 
$a\in$ & $\left[0,0.2\right]$ & $\left(0.2,0.4\right]$ & $\left(0.4,0.6\right]$ & $\left(0.6,0.8\right]$ & $\left(0.8,1\right]$\tabularnewline
\hline 
Counts & 180 & 146 & 325 & 268 & 763\tabularnewline
\hline 
\end{tabular}

\end{center}
\end{table}

\section{Interpretable Models for Driving Behavior}

In this section, we present the models that we investigate. To achieve interpretability, all models take as an input interpretable rules we hand-crafted independently of the dataset.

\subsection{Rules}
\label{sub-sec:rulebooks}

A {\em rule} specifies a desired driving behavior \cite{Censi2020} based on traffic laws, local culture, or consumer expectations. Examples include ``stay in lane" and ``drive under the speed limit". Appendix \ref{our_rules} presents the rules we used in this study.

We evaluate rules over ego's trajectory in a realization. A {\em violation metric} is a function specific to a rule that takes as input a realization, and outputs a \emph{violation score} that captures the degree of violation of the rule by ego's trajectory over the duration of the realization \cite{Censi2020}. For example, %if ego drives faster than the maximum speed limit, 
we can define the violation score for the ``drive under the speed limit" rule as the square ($L2$) norm of how much ego drives faster than the speed limit over the duration of the realization \cite{xiao2021rulebased}.

\begin{figure}[tb]
     \centering
     \begin{subfigure}[b]{0.25\textwidth}
         %\centering
         \includegraphics[width=\textwidth]{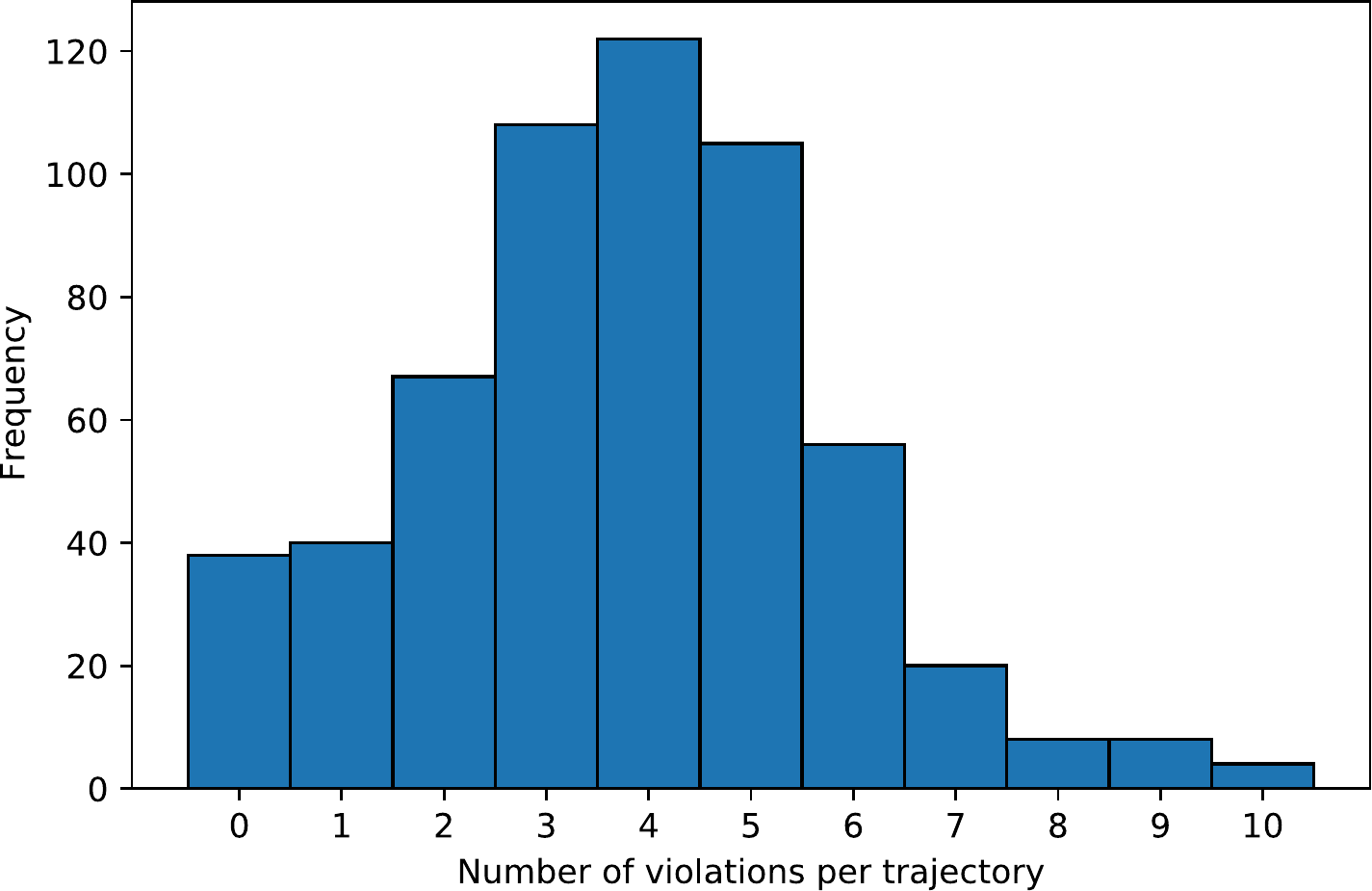}
         \caption{ }
         \label{fig:num_violations_traj}
     \end{subfigure}%
     \begin{subfigure}[b]{0.25\textwidth}
         \centering
         \includegraphics[width=\textwidth]{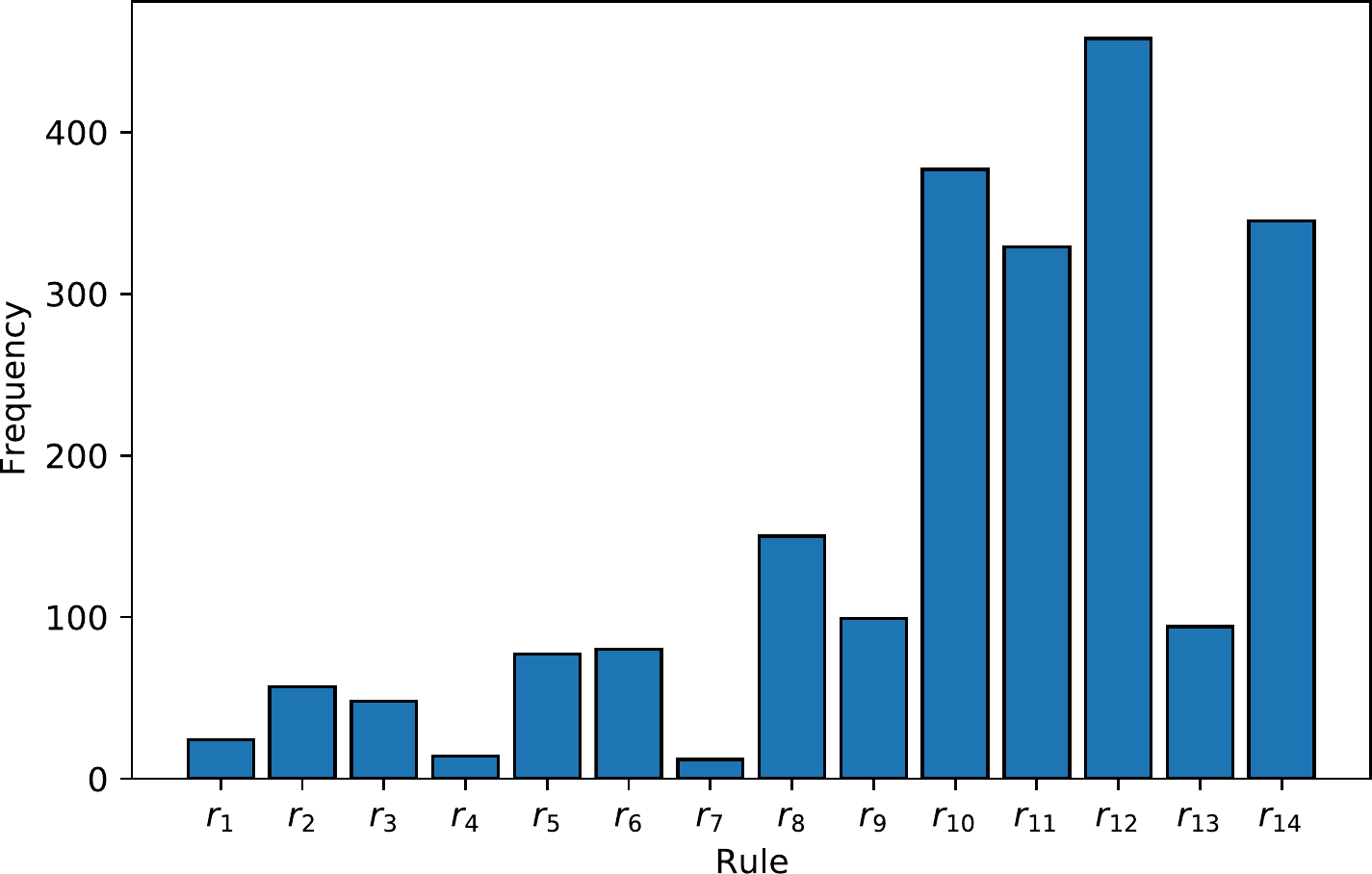}
         \caption{ }
         \label{fig:violation_frequency}
     \end{subfigure}
        \caption{Overview of rule violations in the dataset. \textbf{A}: number of violations per realization. \textbf{B}: violation frequency per rule. Appendix \ref{our_rules} lists and details the rules.}
        \label{fig:dataset_rule_violations}
\end{figure}

Fig.~\ref{fig:dataset_rule_violations} provides an overview of the diversity of violation scores in our dataset. In each realization, ego violated on average 4 rules. In only about 4\% of realizations, ego did not violate any rules. The vehicle clearance rules and the ``Stay in lane" rule were violated most often, whereas collisions and pedestrian clearance rule violations were much less prevalent. Most rule violations are uncorrelated, however some correlations are unavoidable, such as between ``Avoid collisions with Vulnerable Road Users (VRUs)" and ``Maintain clearance with pedestrians off/on the road" rules.

\subsection{Rulebooks}

\begin{figure}[tb]
  \centering
    \includegraphics[width=0.4\textwidth]{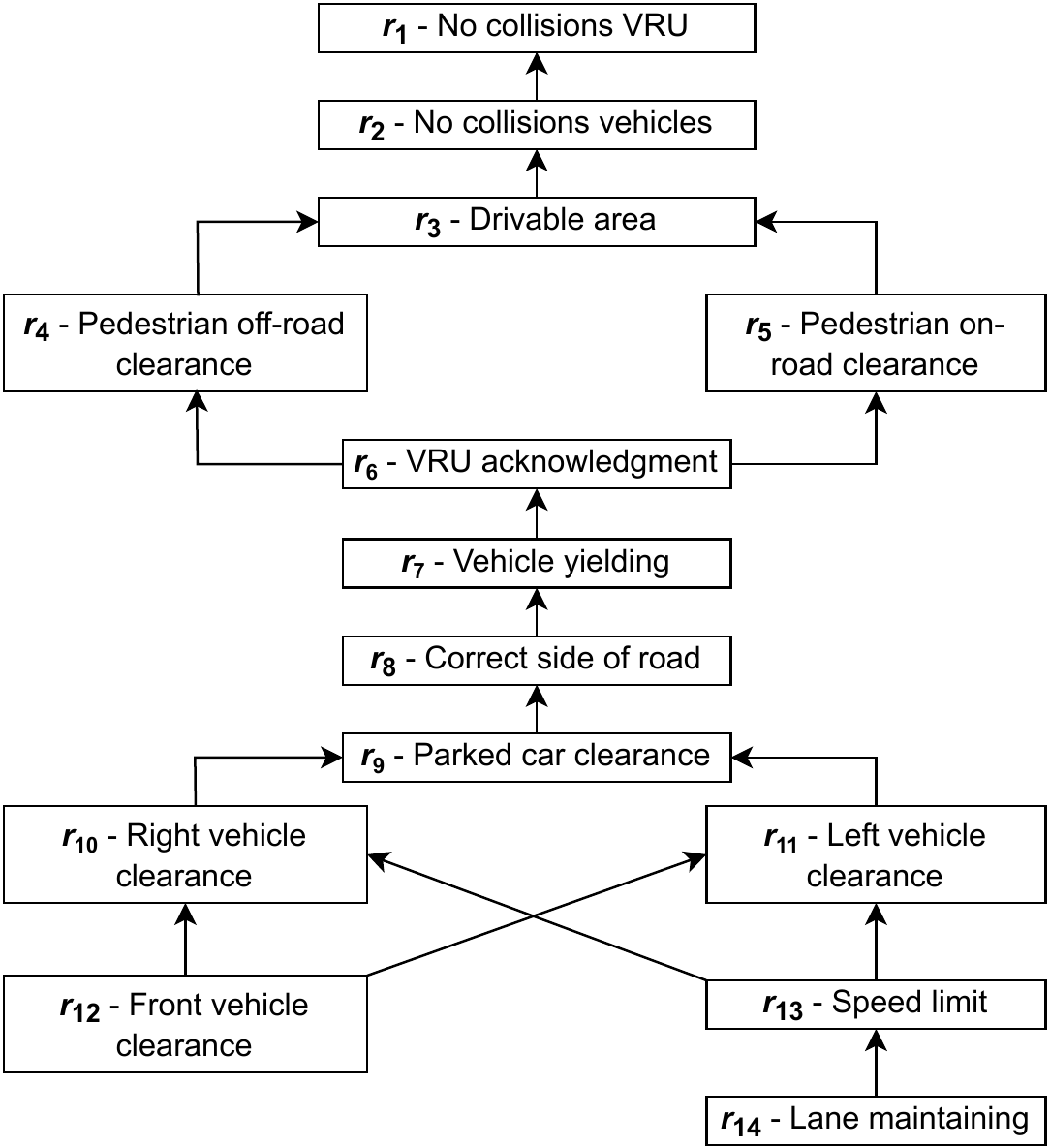}
  \caption{RB: the rulebooks we use to compare realizations. Rules $r_4$ and $r_5$ are incomparable, as are the pairs $(r_{10}, r_{11})$, $(r_{12}, r_{13})$, and $(r_{12}, r_{14})$. Appendix \ref{our_rules} provides full rule names and descriptions.} 
  \label{fig:pre-order}
\end{figure}

A \textit{rulebook} defines relative priorities of rules by imposing a priority structure, which can be a pre-order \cite{Censi2020}, or a total order over equivalence classes \cite{xiao2021rulebased}. 
In this study, we use the pre-order priority structure from Ref. \cite{Censi2020}, in which a rulebook is a tuple $\langle R,\leq\rangle$, where $R$ is a finite set of rules and $\leq$ is a pre-order on $R$. We can represent a rulebook as a directed graph. Each node is a rule and an edge between two rules indicate their relative priorities (see Fig. \ref{fig:pre-order} for an example). 
Formally, a directed edge from $r\in R$ to $r'\in R$ means that $r\leq r'$ ($r' $ has a higher priority than $r$). Note that, using a pre-order, two rules can be in one of three relations: one has a higher priority than the other, 
both have the same priority, or both are incomparable. 
Two rules in a pre-order graph are incomparable if neither is an ancestor of the other. This means that the rulebook does not specify a preference between violating one rule over the other. For example, the rulebook may specify that it is indifferent between violating vehicle clearance to the right or left of ego.

In this study, we use the rulebook presented in Fig. \ref{fig:pre-order}, which we refer to as RB. Subject-matter experts designed RB, without access to the Reasonable Crowd dataset.
We emphasize that the purpose of including RB in this work is not to propose a specific rulebook as \textit{the} correct driving behavior specification, but rather to explore how to assess and potentially improve the performance of \textit{a} rulebook against preference-based data. 

We use Definition 5 of \cite{Censi2020} to rank realization pairs according to RB and violation scores for ego's trajectory in each realization. RB prefers a realization $w_1$ to a realization $w_2$ (i.e., RB considers $w_1$ to be more reasonable than $w_2$) if the highest priority rule that $w_1$ violates has a lower priority than the highest priority rule that $w_2$ violates. If both realization pairs violate the same highest priority rule, then the RB prefers the realization with the smaller violation score. A realization pair is incomparable if the highest priority rules that each one of the realizations violate are incomparable, or if both realizations do not violate any rules.

\subsection{ML Models}

The ML community has developed various interpretable models, such as logistic regression (LR) and decision trees (DT). We overview the ML models that we use in Appendix \ref{our_models}. To get a sense of the performance versus interpretability trade-off, we also considered non-interpretable models, such as a random forest (RF).

To train the parameters of such models, we pose the learning problem as a classification problem. Given a pair of realizations, $w_1$ and $w_2$, the ML model chooses the one which exhibits more reasonable driving behavior. 

We summarize the models' inputs and outputs as follows:
\begin{equation}
f\left(\bm{v}(w_1)-\bm{v}(w_2)\right)=\begin{cases}
+1 & w_{1}\text{ is more reasonable}\\
-1 & \text{otherwise}
\end{cases}
\end{equation}
where $f$ represents the model and $w_1$, $w_2$ are two realization the model compares. $\bm{v}(w)$ contains the violation scores of a realization $w$:
\begin{equation}
\bm{v}=\left(\begin{array}{cccc}
v_{1}\left(w\right) & v_{2}\left(w\right) & ... & v_{14}\left(w\right)\end{array}\right)^{T}\in\mathbb{R}_{\geq0}^{14}
\end{equation}
where for $i=1,2,...,14$ $v_i\left(w\right)\ge 0$ is the violation score of rule $r_i$ for the realization $w$. If $w$ violates $r_i$, then $v_i\left(w\right)>0$ with larger values indicating more severe violations. Otherwise, $v_i\left(w\right)=0$ indicates no violation of $r_i$. Note that for simplicity, we limit the ML model's inputs to be the difference in $w_1$'s and $w_2$'s violation scores.

\section{Experiments}

\subsection{Evaluation setup \label{eval}}
To compare RB (presented in Fig.~\ref{fig:pre-order}) with the ML models, we need to address a specific characteristic of RB: the presence of incomparable rules, and therefore potentially incomparable realizations. With RB, we obtain 118 incomparable realization pairs out of 1682 pairs. Most of these 118 pairs are incomparable due to $r_{10}$ and $r_{11}$ being the highest priority rules that the realizations violate, and other ones are due to realizations not violating any rule (see Fig.~\ref{fig:dataset_rule_violations}A). We show an example of an incomparable pair in Fig. \ref{fig:qualitative_examples}A, where A-1 violates $r_{10}$ more than A-2, but violates $r_{11}$ less.
 
When RB abstains from comparing two realizations, we use a decision tree trained on the data and at most 4 levels deep, which doesn't lower interpretability much because of the intuitive structure of a decision tree. We denote this augmented RB by RB+DT. 

We use classification accuracy as our main metric to measure performance. Accuracy is the percentage of realization pairs $(w_1, w_2)$ correctly classified as $w_1$ is more reasonable than $w_2$ or vice versa. We determine the ground truth labels by using the Bradley Terry model \cite{bt_model} and the annotations to assign a score to each realization. We label a pair $(w_1, w_2)$ as +1 if we score $w_1$ higher than $w_2$, and -1 otherwise.

Since by its nature the dataset is balanced (for every realization pair $(w_1, w_2)$, a pair with the opposite ground-truth label is $(w_2, w_1)$), we argue that the use of accuracy as a metric is reasonable. 

We also provide two more metrics that give a more nuanced view on performance. 

First, we stratify accuracy based on inter-annotator agreement given by Eq. (\ref{eq:agreement_level}) which relates to the difficulty of classifying a particular realization pair. Second, since multiple annotators compared the same realization pair, an annotator doesn't always agree with the ground truth labels. Therefore, we can compare the model to each annotator by determining who agreed more with the ground truth labels of pairs that that annotator annotated. We denote by $L$ the  percentage  of annotators that the model lost to. We  only  consider annotators who annotated at least 10 distinct pairs for this metric.

Because of the limited size of our dataset, we use repeated nested cross-validation (CV) for accurate metric estimation \cite{model_eval}. An outer CV loop divides the data into 5 folds, with each fold used as a test set. An inner CV loop divides the remaining data into 10 folds to determine the ML models' hyperparameters\footnote{Appendix \ref{our_models} overviews the hyperparameters we sweep over.}. We repeat this process 10 times. Note that each of the folds contain mutually exclusive scenarios.

\subsection{Results}
\begin{figure}[tb]
  \centering
    \includegraphics[scale=0.57]{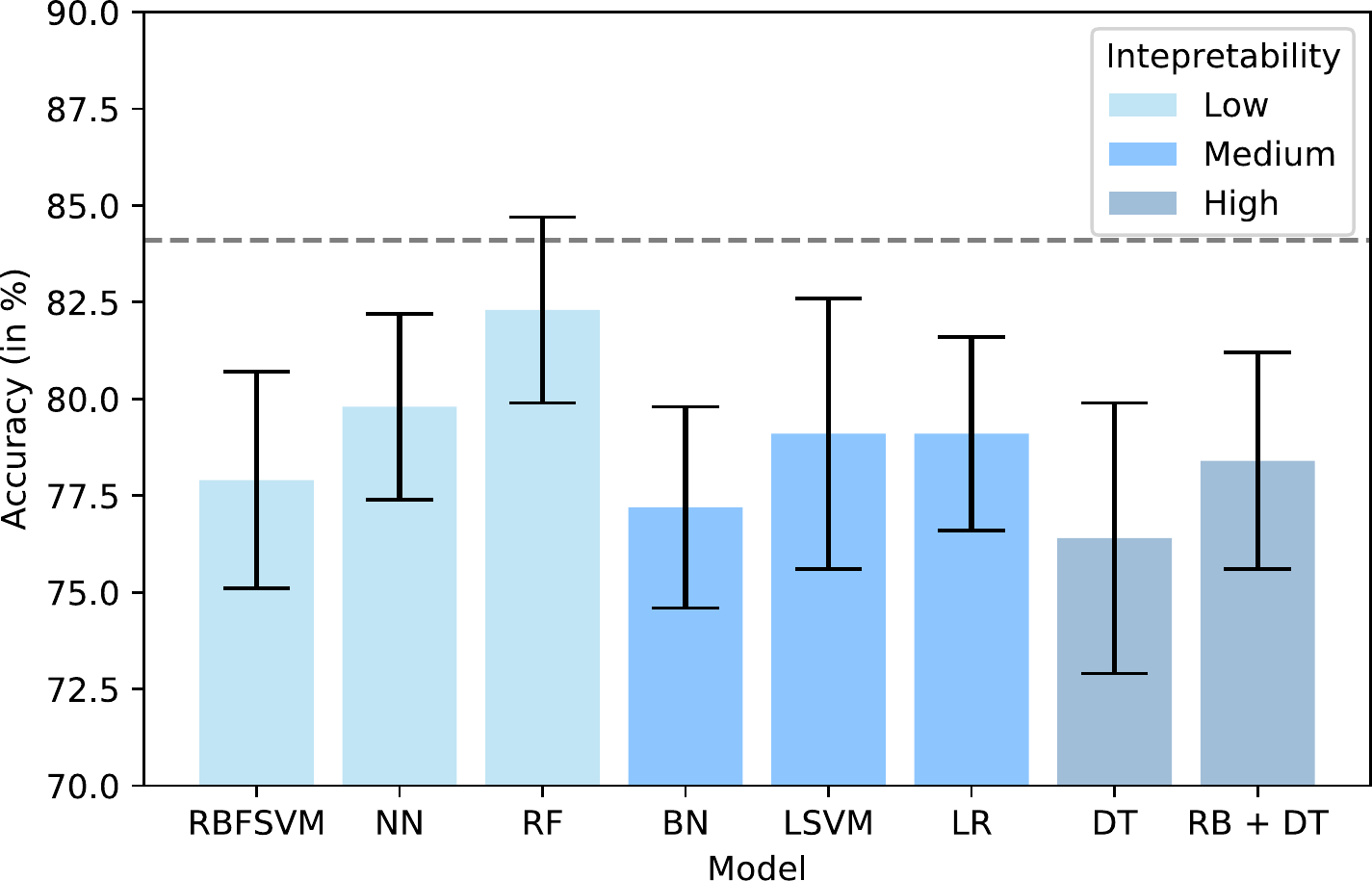}
  \caption{
  Different models' performance on the dataset. Appendix \ref{our_models} defines the ML models RBFSVM, NN, RF, BN, LSVM, LR and DT. Sec. \ref{eval} defines RB+DT. RB+DT is our rulebook augmented with a decision tree. Note that the accuracy of the unaugmented RB on the realization pairs it could compare is 78.9\%. The dashed line is the annotators' median agreement with the ground truth labels (84.1\%).
  }
  \label{main_result}
\end{figure}

\begin{figure}[thpb]
  \centering
    \includegraphics[scale=0.4]{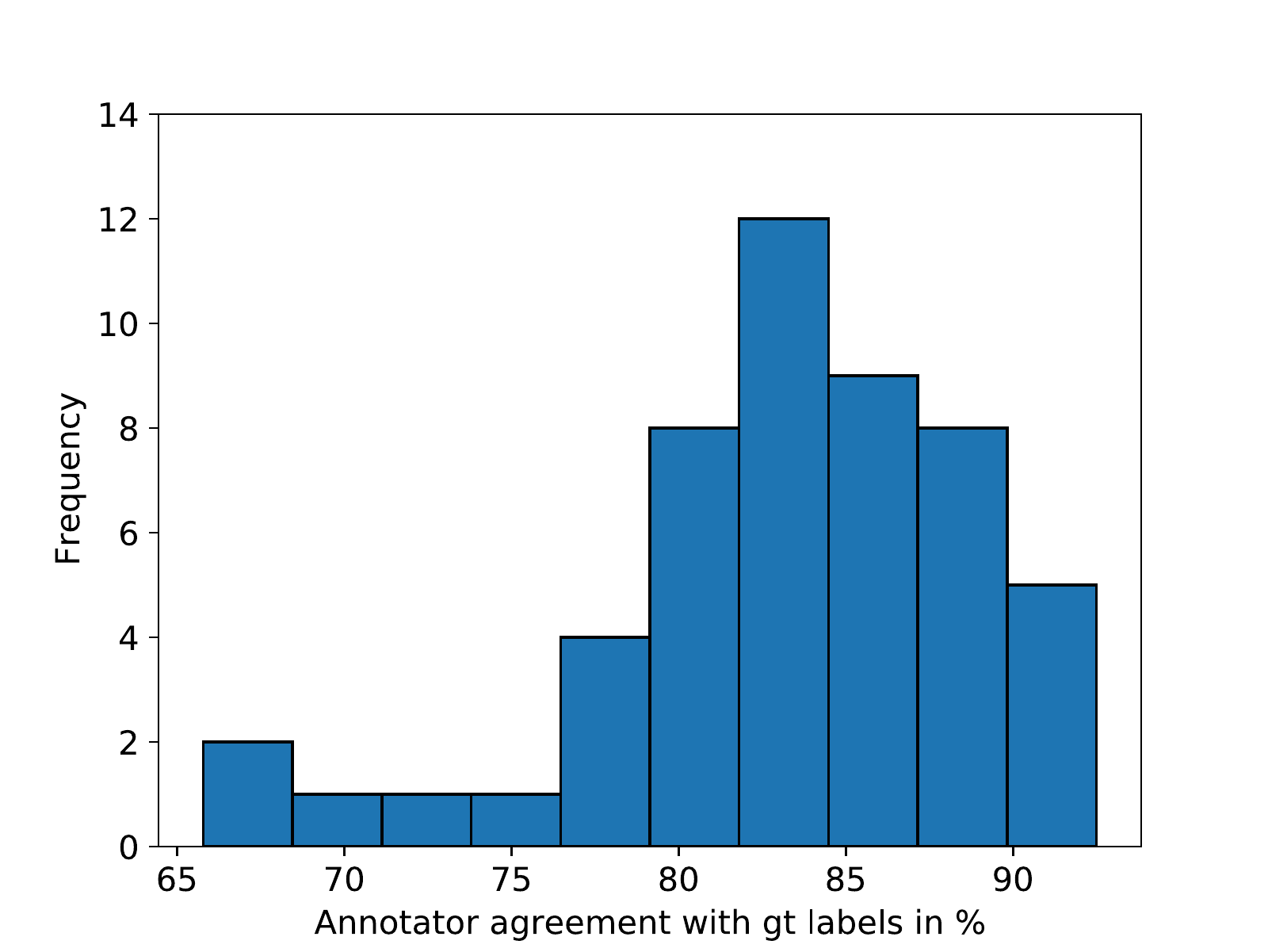}
  \caption{Distribution of annotators' agreement with the ground-truth labels for annotators with at least 10 annotations. The median agreement is 84.1\%}
  \label{ann_acc}
\end{figure}

Fig. \ref{main_result} shows the performance of the different models. Moreover, Table \ref{detailed_metrics} shows detailed metrics for some of the models. To get a sense of what counts as good performance, Fig. \ref{ann_acc} shows the distribution of how well annotators (that annotated at least 10 unique realization pairs) agreed with the ground truth labels. 

Many ML models can output a confidence score for each realization pair $(w_1, w_2)$. We'd expect that this score correlates with the inter-annotator agreement $a(w_1, w_2)$ even though we trained the ML models with only binary labels. Our trained models indeed exhibit such a correlation. The Pearson correlation coefficient between them are $0.66 \pm 0.05$ and $0.73 \pm 0.05$ for LR and RF, respectively, implying that datapoints with higher inter-annotator agreement tend to be further away from a ML model's decision boundary.

\begin{table*}[ht]
\caption{Detailed Metrics. Appendix \ref{our_models} defines RF, LR and DT. Sec. \ref{eval} defines RB+DT and $L$. RB+DT is our rulebook augmented with a decision tree. $L$ is the percentage of annotators a model loses to on ground-truth labels.}
\centering
\begin{tabular}{|c|c||c||c|c|c|c|c|}
\hline 
\multirow{2}{*}{Model } & \multirow{2}{*}{Accuracy (in \%)} & \multirow{2}{*}{$L$ (in \%)} & \multicolumn{5}{c|}{Accuracy (in \%) stratified by $a\in$}\tabularnewline
\cline{4-8} \cline{5-8} \cline{6-8} \cline{7-8} \cline{8-8} 
 &  &  & $\left[0,0.2\right]$ & $\left(0.2,0.4\right]$ & $\left(0.4,0.6\right]$ & $\left(0.6,0.8\right]$ & $\left(0.8,0.1\right]$\tabularnewline
\hline 
\hline 
RF & $82.3\pm2.4$ & $61.4\pm11.2$ & $60.9\pm6.6$  & $65.2\pm8.7$  & $75.7\pm3.9$  & $82.7\pm4.8$  & $93.3\pm1.9$\tabularnewline
\hline 
LR & $79.1\pm2.5$ & $70.5\pm7.4$ & $54.3\pm8.2$  & $65.3\pm9.5$  & $70.6\pm5.3$ & $80.8\pm5.4$  & $90.4\pm3.0$\tabularnewline
\hline 
DT & $76.4\pm3.5$ & $76.5\pm8.5$ & $56.5\pm6.7$ & $65.1\pm8.1$  & $72.2\pm3.8$  & $75.4\pm5.5$  & $86.4\pm3.1$\tabularnewline
\hline 
RB+DT & $78.4\pm2.8$ & $75.8\pm8.3$ & $57.4\pm5.9$ & $66.2\pm7.1$ & $72.4\pm4.5$ & $75.4\pm5.7$ & $86.8\pm3.4$\tabularnewline
\hline 
\end{tabular}
\label{detailed_metrics}
\end{table*}

\subsection{Discussion}

Our results suggest that, even with a limited set of rules, models can adequately capture driving behavior in diverse scenarios with different agent types and road geometries. 

The RF's accuracy is particularly competitive with the annotator's median agreement of $84.1\%$ with the ground truth labels. If we impose interpretability, the performance drops as expected. However, this drop is moderate and we are confident that future work can reduce the gap. 

Among interpretable models, RB performs competitively on the dataset. This is encouraging because RB is the most interpretable model. RB's pre-order priority structure makes it easy to explain why it chose one realization over another. For example, in Fig.~\ref{fig:qualitative_examples}B, RB chose B-1 over B-2 simply because B-2 violates the highest priority rule $r_1$ while B-1 does not.

Furthermore, RB's priority structure places safety above all else (a realization with a collision violation is always worse than a realization with any other violation). There is no such guarantee with ML models whose parameters depend solely on the data. For example, Fig.~\ref{fig:qualitative_examples}B depicts a case where LR and RF chose colliding with a pedestrian as more desirable than colliding with a vehicle. Since safety is the most important requirement for AVs, arguably RB outperforms even the RF in this aspect. 

We could have encouraged the ML models to place more importance on safety by changing the training procedure to for example upweight realization pairs with collisions. However, we note that around 14\% of pairs already contain collisions. How to best incorporate insights from experts and regulatory bodies into ML models is an open question.

Nonetheless, since ML model parameters depend solely on the data, they are easy to scale up. However, it is difficult to scale up the RB priority structure by hand to a wider operating domain than that of our dataset. Indeed, for $n$ rules, we would need to define $n(n-1)/2$ relations, and there are up to $3^{n(n-1)/2}$ possible rulebooks. In addition to traffic law regulations, data is necessary to help narrow this search. 

Data can also inform improvements to individual rules and RB. Through the dataset, we found two potential types of improvements to the rules. First, preference data or human demonstrations can help calibrate free parameters in rules. For example, the data suggested that the clearance threshold chosen for $r_4$ may be too restrictive. In Fig.~\ref{fig:qualitative_examples}C, RB considered an $r_4$ violation for C-2. Inspecting the data, we saw that the crowd did not consider this violation and found ego's lane change in C-1 to avoid an $r_4$ violation to be conservative. We also realized that the dependence of the clearance threshold on speed in $r_5$ may be too lenient. For example, in Fig.~\ref{fig:qualitative_examples}D, D-1 has a larger violation of $r_5$ than D-2.
The annotations suggest a need to increase $r_5$'s dependence on ego's speed to obtain a larger violation of $r_5$ by D-2 in which ego's speed is much higher than in D-1. 

Second, data can help determine what important rules are missing. For example, we identified comfort- and crosswalk-related rules as necessary to specify the desired driving behavior. For example, in Fig.~\ref{fig:qualitative_examples}E, we identified missing yielding rules to give the right of way on the crosswalk to the crossing pedestrian.

The data also indicates a potential limitation with a strict pre-order priority structure for a rulebook. Fig.~\ref{fig:qualitative_examples}C shows a pair of realizations exhibiting a trade-off between violations of $r_{10}$ and $r_8$. F-1 shows a large violation of $r_{10}$ while F-2 shows a minor violation of the higher priority rule $r_8$. Nonetheless, all annotators chose F-2 as being more reasonable. In such trade-offs, a less strict priority structure could ignore a higher priority rule if the violation is below a threshold learned from data.

\section{Conclusions and Future Work}
Setting the desired behavior of AVs is a complex problem that involves numerous stakeholders. In the absence of a widely accepted definition of reasonable driving behavior, data plays an important role in capturing societal preferences. The use of explainable features like rules derived from traffic laws or reasonable expectations can greatly reduce the complexity of the problem and improve the interpretability of models that specify driving behavior. Rules can also simplify AV design and testing by providing machine-readable and objective criteria. However, rules alone are insufficient without a model for resolving trade-offs between them. 

On preference data with a limited operating domain for ego, we found that machine-learned models that leverage a small number of rules achieve a performance comparable to that of human annotators. A rulebook hand-crafted independently of the data performed comparably to all of the interpretable ML models we trained. 

Although our rulebook's hierarchical structure presents some limitations (e.g., the inability to trade off different degrees of violation between rules), it can enforce well-accepted preferences (e.g., prioritizing safety over road etiquette). Moreover, the simplicity of our rulebook's graphical pre-order makes it by far the most interpretable among the models we studied. As the number of rules and amount of data grows, we foresee using combinatoric methods to optimize the rulebook's priority structure.

The data helped us identify improvements in the rule formulations and better understand the strengths and limitations of different  behavior specification models. We hope that this study and the release of the data will stimulate a broader discussion on the desired behavior of AVs and spur innovations in methodologies to specify this behavior. We identify several areas of further work, including methodologies to leverage the data to construct the best-fitting rulebook, comparison of rule-based models against models trained on richer scenario data (e.g., birds-eye view embeddings), and the development of hybrid models that combine the advantages of hierarchical models with more flexible models.

Furthermore, our dataset has room for improvement. It is of a limited size and covers a limited operating domain. For future datasets, we also recommend that annotators comment on their choice after each annotation to increase annotation quality. Moreover, more insightful data can be generated with the use of active learning in the rule violation feature space.

\section*{Acknowledgements}
We thank Eric M. Wolff for many useful discussions, and for help in setting up the data creation pipeline. We thank WeeKiat Ang and Harish Loganathan for help with simulations. We thank Nelly Lyu, Qiang Xu and Ayman Alalao for help with the annotation tool, and data storage and retrieval infrastructure. We thank Nathan Otenti for help with the rulebook implementation.
We thank Shane Chang  and Wei Xiao for their help with early pilots. We thank Juraj Kabzan and Tung Phan-Minh for useful discussions. We thank Emilio Frazzoli and Laura Major for their support.

\bibliographystyle{IEEEtran.bst}

\bibliography{references,Biblio}

\appendices

\section{Rules \label{our_rules}}

\begin{itemize}
\item $r_1$ ``Avoid collisions with VRUs": registers a violation every time there is a collision with a VRU.
\item $r_2$ ``Avoid collisions with vehicles": registers a violation every time there is a collision between ego and other vehicles. The violations of $r_1$ and $r_2$ increase with the severity of impact.
\item $r_3$ ``Stay in the drivable area": registers a violation when ego drives off the road, e.g. into a sidewalk. The further ego creeps out of the road, the higher the violation.
\item $r_4$ ``Maintain clearance with pedestrians off the road": registers a violation when ego comes too close to pedestrians that are off the road surface, e.g. on the sidewalk. The closer ego gets, the higher the violation.
\item $r_5$ ``Maintain clearance with pedestrians on the road": registers a violation when ego comes too close to pedestrians that are on the road surface. The clearance thresholds for $r_4$ and $r_5$ increase with ego's speed.
\item $r_6$ ``Signal intent to maintain clearance with VRU on direct path": registers a violation when ego fails to decelerate enough to acknowledge the presence of a VRU on its path. This rule can be violated when $r_5$ is not. $r_6$ is about perceived safety for other road users. 
%This rule enforces courteous driving behavior, but is allowed to be violated for other, more important rules to be satisfied.
\item $r_7$ ``Yield to vehicles": registers a violation when other vehicles with the right-of-way have to divert from their course because ego failed to yield to them (even if they avoid a collision or near-collision).
\item $r_8$ ``Drive on the correct side of the road": registers a violation when ego drives against the direction of traffic. The longer it is on the wrong side of the road, and the further into the wrong lane, the higher the violation.
\item $r_9$ ``Maintain clearance with parked car": registers a violation when ego drives too close to a parked car. 
\item $r_{10}$ ``Maintain clearance with vehicles on the right": registers a violation when ego drives too close to a vehicle to its right.
\item $r_{11}$ ``Maintain clearance with vehicles on the left": registers a violation when ego drives too close to a vehicle to its left.
\item $r_{12}$ ``Maintain clearance with vehicles on the front": registers a violation when ego drives too close to a vehicle in front of it.
\item $r_{13}$ ``Drive under the speed limit": registers a violation when ego drives too fast. The longer above the speed limit, and the higher the speed, the higher the violation.
\item $r_{14}$ ``Stay in lane": registers a violation when ego crosses a lane boundary (even if legal). The further ego creeps into another lane, the higher the violation.
%This rule captures the fact that changing lanes repeatedly without trying to avoid a safety hazard is not a good driving behavior.
\end{itemize}

\section{ML models \label{our_models}}

\begin{itemize}
\item Logistic Regression (LR): A classifier which uses a weighted linear combination of input features followed by the logistic function. We use the Binary Cross Entropy loss, and skip the bias term in our model. The weight of $L2$ regularization and the number of training iterations are hyperparameters we tune.  
\item Decision Tree (DT): A classification decision tree works by recursively splitting the data into two parts based on each rule. We use the gini criterion. The maximum depth of the tree is a hyperparameter we tune. 
\item Random Forest (RF): An ensemble method where we use multiple decision trees to make predictions and take the majority vote to a produce classification label. Hyperparameters include the maximum depth of a tree and the number of decision trees.
\item Support Vector Machine (SVM): A classifier that tries to find the maximum margin separating hyperplanes between classes. We use the hinge loss and $L2$ regularization. We use the following two types of SVMs:
    \begin{itemize}
        \item Linear SVM (LSVM): This models linear decision boundary in the feature space. The strength of regularization is a hyperparameter used here. 
        \item SVM with Radial Basis Function (RBFSVM): RBF is a Gaussian function applied to the input data. Strength of regularization and scale of the Gaussian function are hyperparameters we tune.
    \end{itemize}
\item Neural Networks (NN): They consist of multiple layers with each layer usually comprising of a linear operation followed by a non-linear activation. We use Binary Cross Entropy loss.  Hyperparameters include number of epochs to train, learning rate, $L2$ regularization strength and number of units in a single hidden layer.
\item Bayesian Networks (BN):
Generative model that represents variables and their conditional dependencies with a directed acyclic graph.
A BN classifier estimates the probability of selecting one of the realizations $w_1$ or $w_2$ given, for each rule, a binary specification of which realization violated the rule more. We used a one-dependence Bayesian classifier using Chow-Liu's algorithm \cite{friedman1997bayesian}.

\end{itemize}

\end{document}